\documentclass[a4paper]{svmult}

\usepackage{type1cm}        
\usepackage{makeidx}         
\usepackage{graphicx}        
\usepackage{multicol}        
\usepackage[bottom]{footmisc}
\usepackage{newtxtext}       %
\usepackage{newtxmath}       
\usepackage{wrapfig}
\usepackage[justification=centering]{caption}
\usepackage{lipsum}
\usepackage{hyperref}
\usepackage{authblk}

\makeindex             
                       
\usepackage{microtype}

\usepackage[font=small]{caption}
\usepackage{array}
\newcolumntype{P}[1]{>{\centering\arraybackslash}p{#1}}

\begin{document}

\title*{Digger Finger: GelSight Tactile Sensor for Object Identification Inside Granular Media}
\titlerunning{Digger Finger}
\author{Radhen Patel, Rui Ouyang, Branden Romero, Edward Adelson \\
Massachusetts Institute of Technology \\ < radhen, brromero > $@$mit.edu, nouyang@alum.mit.edu, adelson$@$csail.mit.edu
}
\authorrunning{R. Patel et al.} 


\maketitle

\vspace{-1.5cm}
\section{INTRODUCTION} 
\label{sec:intro} 
\vspace{-0.25cm}

\begin{wrapfigure}[18]{r}{0.55\textwidth}
\centering
\includegraphics[width=0.55\textwidth, trim=0 0 0 0, clip, keepaspectratio]{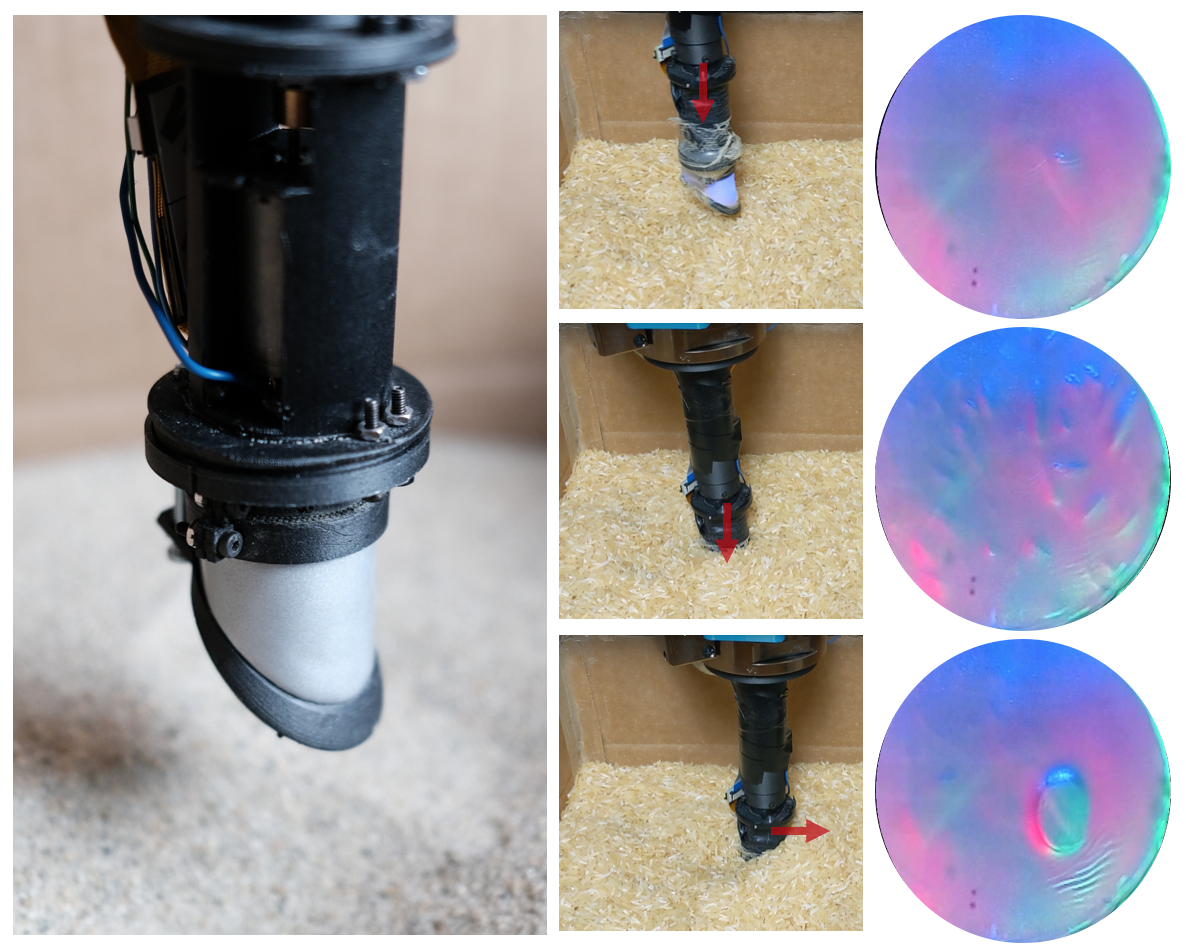}
\caption{\textit{Left:} Digger Finger. \textit{Middle:} Penetration motion. \textit{Right:} Tactile data showing zero contact, granular media (rice), and object contact}.
\label{real}
\end{wrapfigure}

Imagine you are at the beach with a metal detector which goes off, and you stick your hands in the sand to find the metal object. Even though the granular media (sand) is constantly affecting your sense of touch on your fingers and palm, its acuity combined with your cognition enables you to easily find buried objects. Manipulating granular media, just like manipulating rigid or deformable objects, comes very naturally to us. For robots, however, it remains a challenging task. Research in robotic manipulation has focused on rigid objects more than deformable objects and granular media. One reason is due to the difficulty of modelling the complex dynamics of the latter two. Another reason is that the perceptual understanding of the latter two via tactile based hardware devices and algorithms, compared to vision-based methods, is poor. Yet, when dealing with physical interactions, tactile sensation can be more critical than visual information. Due to these compounding limitations, robotic manipulation of deformable objects and granular media remains poorly explored. With this motivation, we take on the challenge of using touch feedback to search for objects buried in granular media. A robot with such capabilities can prove useful in areas such as deep sea exploration \cite{nadeautactile}, mining, excavation, decommissioning explosive ordinates \cite{burtness2011evaluation, noauthor_t4_2019}, agricultural robotics and other areas where task dependant information can be occluded. In this paper we present an early prototype of the Digger Finger (Figure \ref{real}) that is designed to easily penetrate granular media and is equipped with the GelSight sensor \cite{Johnson2009RetrographicSF, johnson2011microgeometry}.

We begin by providing a brief overview of the related work in the domain of robotic manipulation for granular media. We then present our technical approach in section \ref{approach} starting with the design of the sensor in section \ref{sec:sensordesign} and then the manufacturing process in section \ref{sec:manu}. Section \ref{experiments} describes the two experimental procedures we used to evaluate the performance of the Digger Finger in searching object buried inside granular media. Section \ref{sec:fluid} illustrates the ability of the Digger Finger to fluidize granular media during penetration. Section \ref{sec:object} illustrates the ability of the Digger Finger to identify objects that are buried inside granular media. Finally, in section \ref{sec:insights} we provide our concluding remarks. 

Robotics research related to granular media has been fall primarily within the scope of automated operation of construction equipment such as scooping \cite{sarata2004trajectory}, legged locomotion \cite{li2013terradynamics, hauser2016friction}, gripper design\cite{brown2010universal}, manipulators \cite{cheng2012design}, haptic displays \cite{brown2020soft, stanley2013haptic} and in robotic pouring tasks \cite{yamaguchi2015pouring, matl2020inferring}, to name a few. In contrast, work on robotic manipulation of and within granular media has only recently begun receiving attention from the research community. In \cite{schenck2017learning}, the authors teach a robot how to scoop and dig into a pile of beans by learning the dynamics of the media from visual data. Building on \cite{schenck2017learning}, the authors in \cite{clarke2019robot} and \cite{clarke2018learning} learn to use tactile, visual and auditory feedback to estimate the flow and amount of granular materials during scooping and pouring tasks respectively. Regarding object identification in granular media in particular, two non-destructive methods that are popular for finding buried objects are Ground Penetrating Radars (GPR) and ultrasonic vibrations \cite{travassos2020artificial, martin2002ultrasonic}. These methods, though reliable, are only good at estimating rough geometry and approximate location of the buried objects. Closest to our work is \cite{jia2017multimodal} and \cite{syrymova2020vibro}. In \cite{jia2017multimodal} the authors use the BioTac sensors from SynTouch Inc. on a three fingered Barrett hand to detect contact with a cylinder fixed inside a bed of granular media. They take advantage of multi-modal sensory data to classify contact and no-contact events. Similarly in \cite{syrymova2020vibro} the authors again only classify the presence or absence of an object inside granular media. Unlike \cite{jia2017multimodal, syrymova2020vibro}, we use the high resolution data from the Digger Finger to identify different objects inside granular media. Also, our design enables deeper penetration in granular media with the help of mechanical vibrations.

\section{TECHNICAL APPROACH} \label{approach}
\vspace{-0.25cm}

\subsection{Sensor Design} \label{sec:sensordesign} \vspace{-0.5cm}
The three main goals of our design were, ($i$) enable the sensor to easily penetrate the granular media, ($ii$) provide rich tactile sensing to identify objects that are buried inside granular media. ($iii$) achieve human finger like form factor so that the sensor can easily be fitted on existing robot hands.
\begin{wrapfigure}[35]{R}{0.35\textwidth}
\centering
\includegraphics[width=0.35\textwidth, keepaspectratio, trim=0cm 0.25cm 0.5cm 0.5cm, clip]{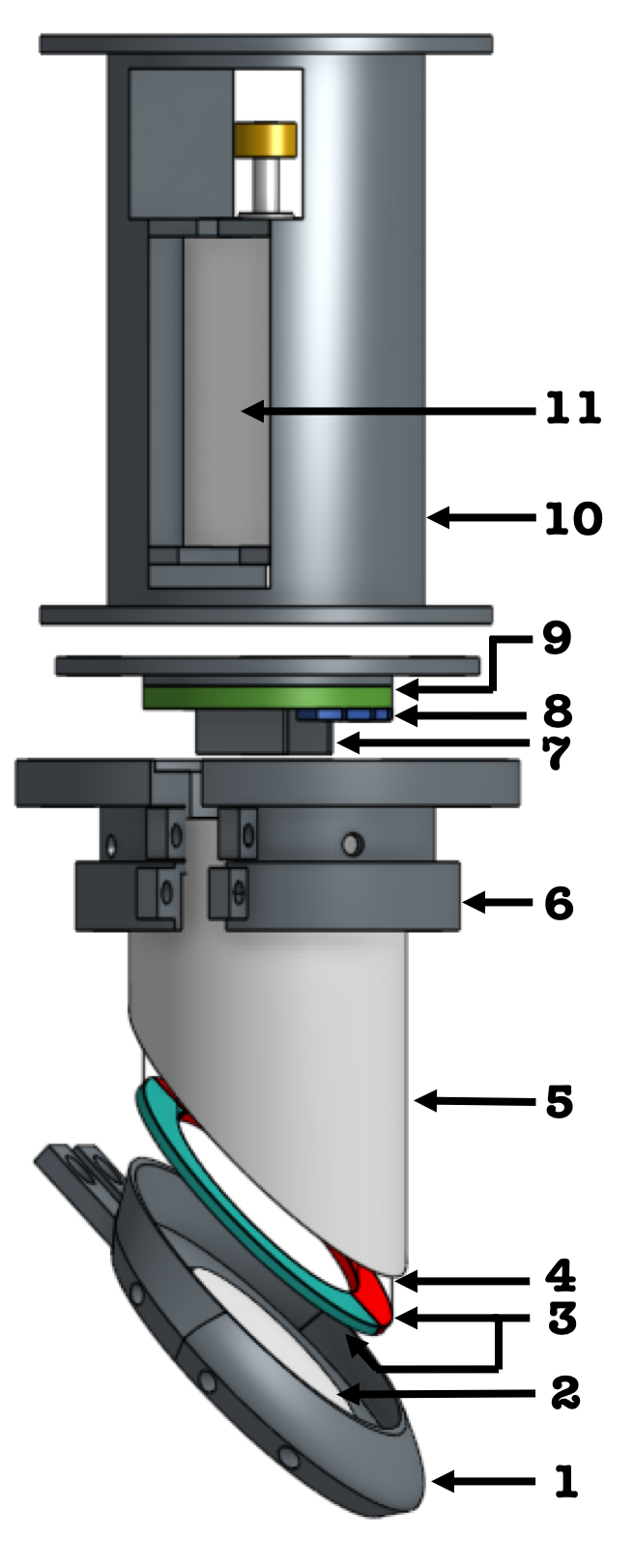}
\caption{Exploded view of the Digger Finger. Numbered arrows are 1. Bottom housing, 2. Mirror 3, Fluorescent paint, 4. Clear acrylic tube, 5. Gel, 6. Top housing, 7. Camera housing, 8. Blue LEDs, 9. PCB, 10. Vibrator housing, 11. Vibrator motor}.
\label{CAD}
\end{wrapfigure}  
The GelSight \cite{Johnson2009RetrographicSF, johnson2011microgeometry} is one type of vision based tactile sensor \cite{abad2020visuotactile, shimonomura2019tactile}. It consists of a camera that observes a directionally illuminated clear elastomer gel coated in a Lambertian or semi-specular membrane. Objects pressed into the gel deform the surface and the illumination allows estimation of surface normal along the deformation. The 3D geometry of the deformation can then be recovered using photometric stereo. Previous GelSight sensors had flat sensing surfaces \cite{Yuan2017GelSightHR, Donlon2018GelSlimAH}, which restricted their integration with manipulators to parallel grippers for planar contact interactions. The four major modifications that we make to the previous GelSight sensors to make it suitable for searching for objects in granular media are ($i$) sensor shape, ($ii$) illumination source, ($iii$) addition of mechanical vibration and ($iv$) gel. Following are the details on the same. 

First, our design adds a pointed tip to the prior work on creating curved GelSight sensors \cite{Romero2020SoftRH}. Second, we replace two color LEDs (green and red) with fluorescent acrylic paint from Liquitex. This allows for a simple and compact design especially at the tip of the Digger Finger which has to face the brunt of digging. We use six blue color LEDs (3528) from  Chazon to excite the fluorescent paints from the top of the Digger Finger (ref. Fig. \ref{CAD}). To excite the red and green color paint to the required intensity we shine an excessive amount of blue light into the Digger Finger. This causes the image to be dominated by the blue spectrum compared to the red and green parts. To compensate this we use a small piece of yellow filter that we put on top of the image sensor of the camera by placing the filter inside the lens assembly. We also experimentally set the white balance setting of the camera to suite our needs. The camera that we use is from Arducam (SKU: B006603) which we interface with using Raspberry Pi 4. Third, this new design aids digging by fluidizing granular media using vibrations. For this we mount a high speed micro vibration motor (6$-$12\,$V$, 18000\,$rpm$) on to the Digger Finger. Fourth, we replace the silicone gel used in the previous GelSight sensors with a 3\,$mm$ wide and 1.5\,$mm$ thick double sided, transparent polyurethane tape (ZSHK Happy Cover HC06). We also keep two things from previous GelSight designs. First, we use a mirror as did \cite{Donlon2018GelSlimAH} to have a camera view that is perpendicular to the gel sensing surface, rather than looking at the surface at an angle. This allows us to have a slim device without worrying about self-occlusions of the sensing surface when contact is made. Second, we use the concept of light piping illustrated in \cite{Romero2020SoftRH} to have light rays pass across a curved surface with negligible loss in light intensity.

\subsection{Manufacturing Process} \label{sec:manu} \vspace{-0.5cm}
The manufacturing process of the Digger Finger consists of several simple rapid prototyping techniques including 3D printing, laser cutting and spray painting, to name a few. The Digger Finger consists of a sensing module and a vibration module. The Digger Finger's sensing module's core is shaped like a cylindrical wedge. We use a piece of optically clear acrylic tube with inner diameter of 16\,$mm$ and  outer diameter of 22\,$mm$ (see Fig. \ref{CAD}). The tube is first diagonally cut using a band saw. The top and bottom surfaces of the cut tube are then sanded and polished until optically clear. The bottom surface of the tube is then painted with red and green fluorescent paint. Afterwards, we design and 3D print two custom housing (top and bottom) to house the necessary components of the Digger Finger (cameras, LEDs, mirror, gel boundaries). A thin piece of mirror is cut in shape of an ellipse using a laser cutter and attached to the center of the bottom housing using VHB double sided tape. Blue colored LEDs are soldered onto a custom designed printed circuit board (PCB) that is milled from a copper plate. This PCB is friction fit to the top housing along with the camera. The polyurethane double-sided tape that we use as the gel is brushed with aluminum flake powder, and then over coated with a thin layer of thermoplastic polyurethane (TPU) dissolved in a solvent. The gel is left to rest until the coating is dry, after which the uncoated side is applied across the clear acrylic tube. With the fluorescent paint and gel on the acrylic tube, we then assemble the top and bottom housing with the acrylic tube. This finishes the complete assembly of the sensing module of the Digger Finger. The vibration module is a 3D printed housing for the vibrator motor which is also cylindrical in shape. This module gets attached on top of the sensing module using screws and nuts. 

\section{EXPERIMENTS AND RESULTS}
\label{experiments}
\vspace{-0.25cm}

\subsection{Fluidizing Granular Media} \label{sec:fluid} \vspace{-0.5cm}
The goal of this experiment is to study the effect of mechanical vibrations on the jamming of granular media. An immersed intruder moving in a granular media will experience strong forces because of the particle jamming effect. These forces are a function of several geometrical and material properties of the intruder and the granular media particles. It is well studied that the force chains formed among the granular particles during jamming can be weakened and the granular media fluidized by blowing air \cite{brzinski2010characterization, naclerio2018soft} or by mechanical vibrations \cite{firstbrook2017experimental, texier2017low}. Following \cite{texier2017low} we use a vibrator motor to study the relation between the force acting on the Digger Finger and its penetration depth in granular media, in the presence and absence of mechanical vibrations.

\begin{figure}[!h]
\centering
\includegraphics[width=\textwidth, keepaspectratio, trim=0cm 0cm 0cm 0cm, clip]{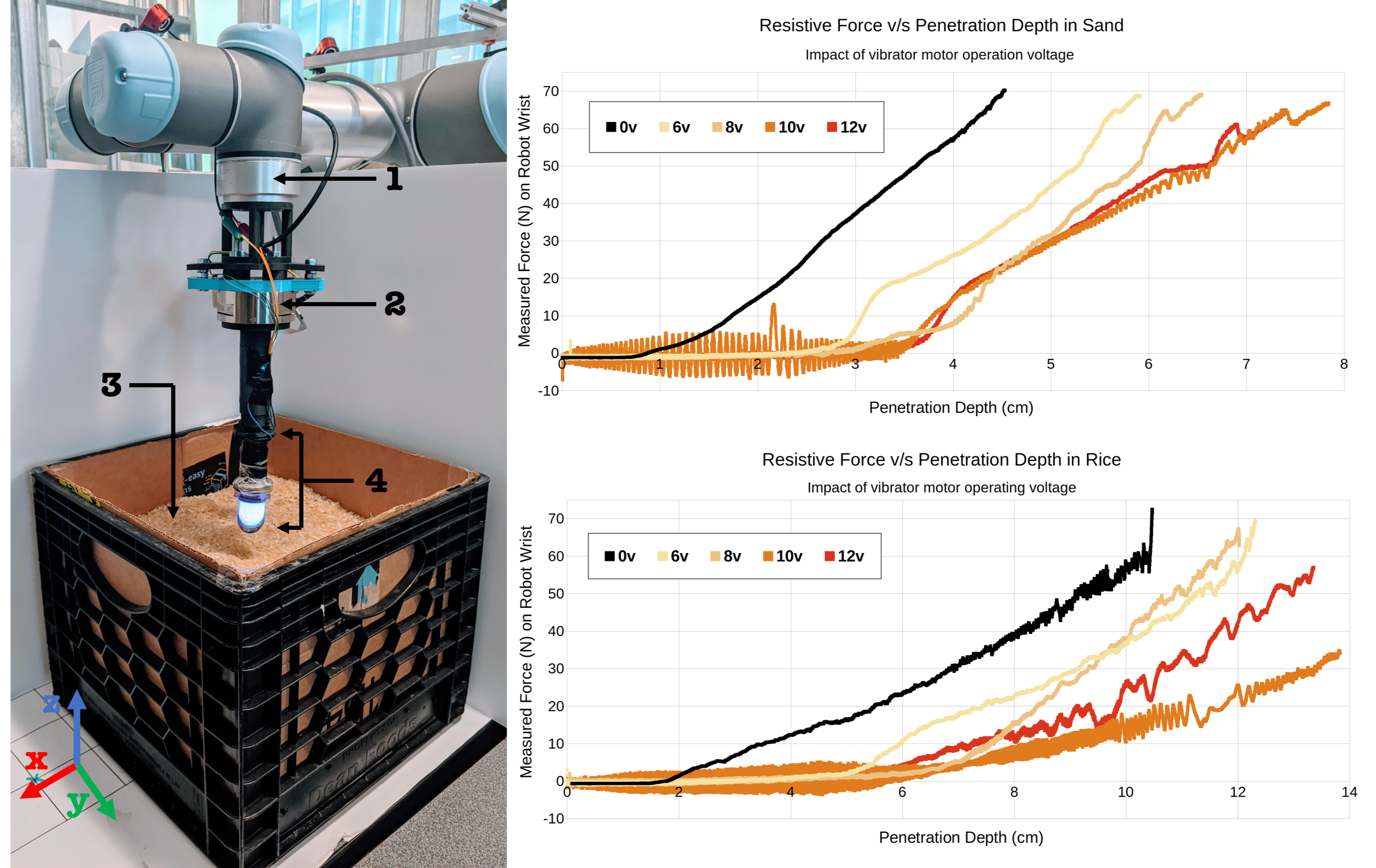}
\caption{\textit{Left:} Experimental setup. Arrows show 1. UR5 robot arm, 2. F/T sensor, 3. Granular media (rice), 4. Digger Finger \textit{Right:} As vibrator motor operating voltage increases, the force (measured by F/T sensor) required to move a given vertical distance in the granular media decreases.}
\label{exp_1}
\end{figure}

The experimental setup as shown in Figure \ref{exp_1} (left) consists of a robot arm (UR5) from Universal Robots Inc. The Digger Finger is coupled to the robot wrist using a 3D printed part. We use two types of granular media in our experiments, sand and rice. We choose these two particular media because of two very distinct effects they have on the object identification task, explained in the next section. For the sake of this experiment it is more important to note that both these media have bulk densities equal to 1578.56\,$kg/m^3$ and 941.48\,$kg/m^3$ respectively, resulting in different force versus distance relationships. Both the granular media are placed in a plastic container (12$x$12$x$11\,$inches$) up to 6\,$inch$ deep. We choose these container dimensions to reduce the \textit{edge effect} during penetration \cite{seguin2008influence}. We attach a 6-axis force torque sensor (ATI Gamma) at the wrist of the robot arm to measure the ground truth forces acting on the Digger Finger during penetration. For each experimental trial the robot arm is first positioned such that the tip of the Digger Finger touches the surface of the granular media. The robot arm is then commanded to vertically penetrate the granular media at a speed of 2\,$mm/s$. We keep the velocity of penetration fixed for all trials. The robot arm is moved until it reaches the safety limit of its joints and stalls. We experimentally find that this distance, both in the case of sand and rice, is less than the depth of the container. In the first trial the vibrator motor is off. For each of the next trials we increase the operating voltage of the vibrator motor by 2\,$V$, starting from 6\,$V$ and going up to 12\,$V$, thereby increasing the vibrating frequency and amplitude of oscillation of the motor. Prior to each trial, the state of the granular media is reset by vigorously shaking the container and shoveling the granular media with a tool. For each trial the 3D position of the robot wrist and force from the ATI Gamma sensor are synchronously recorded. The force in z-axis is first smoothed using an exponential filter ($\alpha=0.1$) and is then plotted against the $z$ position vector of the robot arm for each trial as shown in Figure \ref{exp_1} (right). It is evident from the figure that when the vibrator motor is off i.e. at 0\,$V$, the force on the Digger Finger begins to increase earlier in contrast to when the vibrations are on. This early rise in the force is approximately 2\,$cm$ in the sand case and 4\,$cm$ in the rice case. Moreover, in the cases when the motor is on, all the curves are slightly less steep resulting in almost double the penetration at stall as compared to the case when the motor is off.

\begin{wraptable}[11]{r}{5.5cm}
\centering
\begin{tabular}{|P{1.75cm}|P{1.75cm}|P{1.75cm}|P{1.75cm}|}
	\hline
	Motor operating voltage ($V$) & Frequency ($Hz$)  & Acceleration amplitude ($m/s^2$) \\
	\hline\hline
	6 & 156 & 9.6  \\
	\hline
	8 & 189 & 19.8 \\
	\hline
	10 & 213 & 23.6 \\
	\hline
	12 & 172 & 14.7 \\
	\hline
\end{tabular}
\caption{  Vibrating frequency and acceleration amplitude of the Digger Finger tip as a function of vibrator motor voltage.}
\label{t1}
\end{wraptable}

We quantify the vibrations from the vibrator motor at 6\,$V$, 8\,$V$, 10\,$V$ \& 12\,$V$ voltages in terms of frequency and acceleration amplitude using a 3-axis accelerometer. As the motor is rated to run at 18000\,$rpm$ at 12\,$V$, theoretically the motor cannot vibrate more than 300\,$Hz$. So we use an accelerometer present in LSM9DS1 sensor chip \footnote{\url{https://www.sparkfun.com/products/13944}} that we can sample at a maximum frequency of 500\,$Hz$. The board is attached to the tip of Digger Finger using a double sided tape. The vibrator motor is run at the above specified four operating voltages and data from the accelerometer is recorded for five seconds. The fundamental frequency of the vibrations is found by running fast Fourier transform on the data. The average of the fundamental frequency over two trials for the four operating voltage is shown in Table \ref{t1}. The drop in the frequency and acceleration amplitude going from 10\,$V$ to 12\,$V$ is attributed to the resonance in the mechanical vibrations that we observe at 10\,$V$. Due to this resonance the Digger Finger vibrates with the largest amplitude at 10\,$V$. This is clearly visible in Figure \ref{exp_1}. While all the others curves are smoothed out with the same smoothing constant (i.e. $\alpha=0.1$), the curve at 10\,$V$ still shows periodic oscillation at certain sections of the curve, both for sand and rice, indicating the need for an $\alpha<0.1$ to obtain a smooth curve like the rest. We further verify the vibration frequency using the accelerometer on a smart phone using two Android  software applications. One (iDynamics \footnote{\url{https://www.bauing.uni-kl.de/en/sdt/idynamics/}}) analyzes mechanical vibration and the other (Spectroid \footnote{\url{https://download.cnet.com/Spectroid/3000-20432_4-77833231.html}}) audio. We find that the fundamental frequency values for the four operating voltages range between 160 - 210\,$Hz$. The above observations clearly indicate that the Digger Finger is able to fluidize densely packed granular media with vibrations of approximately 150 - 200\,$Hz$ and 10 - 24\,$m/s^2$ acceleration amplitude, resulting in deeper penetration.

\subsection{Object Identification} \label{sec:object} \vspace{-0.5cm}

\begin{figure}[!h]
\centering
\includegraphics[width=\textwidth, keepaspectratio, trim=0cm 0cm 0cm 0cm, clip]{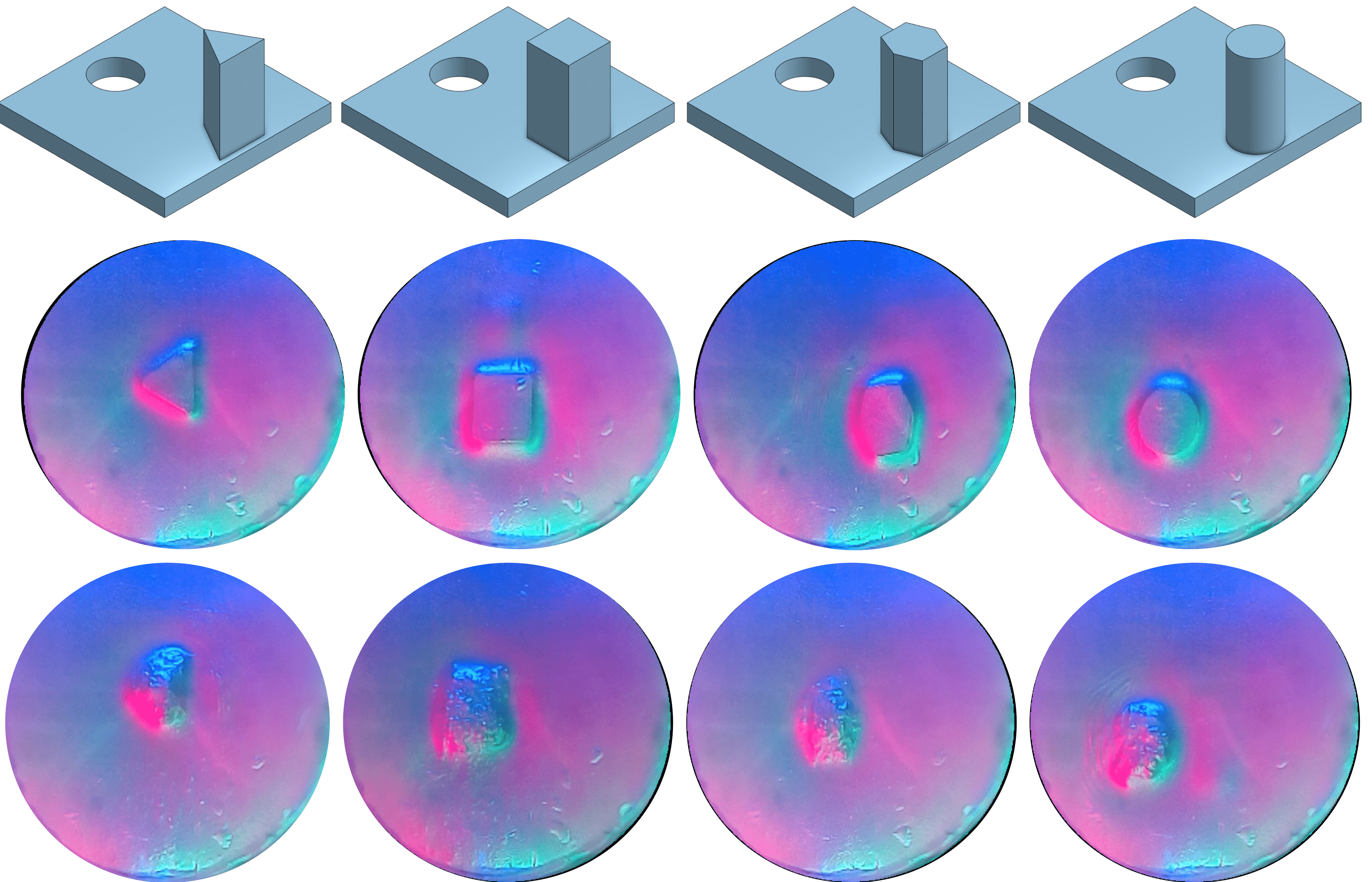}
\caption{\textit{Top row:} Design of buried 3D printed objects. \textit{Middle row:} Digger Finger image showing imprint of object shapes in absence of any granular media. \textit{Bottom row} Imprints of shapes in the presence of granular media (sand).}
\label{shapes}
\end{figure}

This experiment aims to study the performance of the Digger Finger in identifying objects in different granular media. For simplicity we narrow down the problem to classify four simple shapes i.e. triangle, square, hexagon and circle as shown in Figure \ref{shapes}, in two different types of granular media i.e. sand and rice. We choose rice and sand because both media have exhibit very different behavior in the way their grains interact with the Digger Finger and the object shapes we are interested in identifying. Through ad hoc experiments with different granular media (washed sand, chia seeds, lentils, and mung beans) we observed that there are three distinct phenomenona that occur when a Digger Finger immersed in granular media approaches a stationary object, depending upon the geometry and material properties of the grains as well as the Digger Finger and the object. The first occurs when very small grains such sand get permanently stuck between the Digger Finger and the object. The second occurs with grains (in our case rice) that are similar in size to the buried object, in which case the grains can completely block the object from coming in contact with the Digger Finger. The only way to come in contact with the object is to vibrate or twist the Digger Finger to push the grains to the side (refer to supplementary video \footnote{\url{https://sites.google.com/view/diggerfinger}}). We found twisting to be the most efficient way of pushing rice grains to the sides. Third case is when the material properties of the grains e.g. mung beans and lentils are such that they become slippery and do not get stuck between the Digger Finger and object. Based on this ad hoc experiment we narrow down the choice of granular medias to sand and rice. Figure \ref{shapes} shows the different 3D printed objects that we make use of to collect data for this experiment and their corresponding imprints on the Digger Finger as seen from the camera in the form of RGB images.   

For data collection we manually press the 3D printed objects on the Digger Finger and collect around 3000 images for each object shape. We repeat this procedure in a container filled with sand to collect images for cases when the sand grains get stuck between the Digger Finger. As shown in Figure \ref{shapes}, the sand grains distort the boundaries of the object shapes, potentially making them ambiguous. These two rounds of data collection result in a total of eight classes: four classes with just the object shapes and four with sand obscuring the objects. We add a ninth class for the  zero contact case. The dataset for the first eight classes are cleaned by manually removing images where the object is making no or partial contact with the Digger Finger. This data cleaning process leaves us with around 1500 images for each of the first eight classes. 

\begin{wrapfigure}[18]{r}{0.6\textwidth}
\centering
\includegraphics[width=0.6\textwidth, keepaspectratio, trim=0cm 0cm 0cm 0cm, clip]{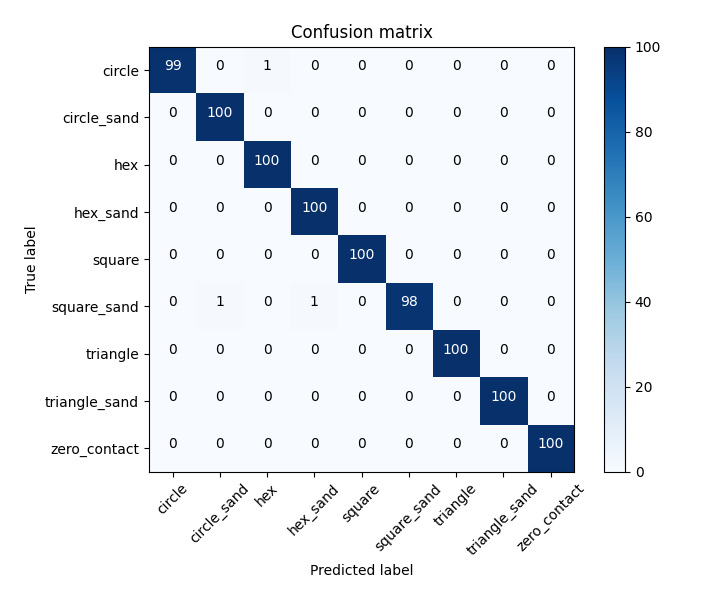}
\caption{Confusion matrix for object identification task}.
\label{confusion}
\end{wrapfigure}

For classification we rely on convolution neural networks because of their growing popularity for image analysis. In particular we train a residual neural network \cite{he2016deep} (ResNet50) on our data set by performing a popular technique in machine learning called transfer learning. Transfer learning focuses on storing knowledge gained while solving one problem and applying it to a different but related problem. For example, ResNet50 network is often first trained to classify real world images from ImageNet \footnote{\url{http://www.image-net.org/}} data set. We use this pre-trained network as the starting point and re-train it on our own object shape data set. We do this by only training the last convolutional block of the network and an uninitialized fully connected layers (classifier block) of size 128 at the end of the network. We split the 1500 images of each class into training (1200), validation (200) and testing (100) data sets. We augment the training data by randomly cropping and rotating the images. We also add Gaussian noise to each color channel. The network is trained for 10 epochs with a learning rate scheduler and batch size of 64, at the end of which it attains 99$\%$ training 98\% validation accuracy. The confusion matrix calculated on the test images is shown in Figure \ref{confusion}.

\section{CONCLUSION AND FUTURE WORK}
\label{sec:insights}
\vspace{-0.25cm}

Identifying objects buried in granular media using tactile sensors is a challenging task. In our experiments we identify several difficulties. First, it is difficult to actually reach the object because the granular media will start to jam and prevent downward movement. Second, even when the tactile sensor can reach the object, the granular media particles tend to get stuck between the sensor and object, distorting the actual shape of the object. To tackle these challenges we present a novel tactile sensor that we call Digger Finger. We build on previous GelSight tactile sensor designs and introduce several innovations, including the use of red and green fluorescent paint and polyurethane tape as the gel, to design a compact wedge-shaped sensor. We design the Digger Finger to fluidize granular media during penetration using mechanical vibrations. We use the high resolution tactile sensing provided by the Digger Finger to successfully identify different object shapes even when distorted by granular media particles. 

For fluidizing granular media, we have only presented results for vertical penetration, but moving horizontally in granular media is also increasingly challenging at greater depths. Further experiments are required to understand the nature of vibration and robot arm motion needed to fluidize granular media while moving horizontally. For object identification a major issue is that over time the paint on the gel incurs wear causing noticeable artifacts in the RGB sensor image data. The tactile data after wear looks different than the data used to train the neural network for object identification. This causes the network to make false positive predictions, in particular for the zero contact class. The network may also make false predictions when the granular media touches the Digger Finger during free motion or when the gel wrinkles during penetration. Ideally, the network should be able predict the type of the granular media and also robustly predict zero contact. We therefore plan to train the network on additional classes of granular media and also experiment with better data augmentation techniques to make the network more robust to gel artifacts. We also plan to calibrate the Digger Finger in order to construct 3D geometry data from the raw RGB image data. In particular this would let us use algorithms trained on simulated 3D geometry data, as opposed to trying to simulate raw RGB sensor image which requires modelling the exact illumination of the Digger Finger. Using 3D geometry data would also allow object detection algorithms to better transfer to real world tactile data from multiple Digger Finger units. 

We believe that such a tactile sensor paves the way for the robotic manipulation within and of granular media, whether for everyday tasks such as scooping rice or litter, or for more industrial applications such as finding and inspecting buried cables and other objects.

\section{ACKNOWLEDGEMENTS}
\vspace{-0.25cm}
This research was supported by the Toyota Research Institute, the Office of Naval Research (ONR) [N00014-18-1-2815], and the GentleMAN project of the Norwegian Research Council. We would like to thank Achu Wilson, Shaoxiong Wang, Sandra Liu, Yu She, Filipe Veiga, and Megha Tippur for insightful discussions. We also thank Siyuan Dong, Daolin Ma and Alberto Rodriguez for lending us the force-torque sensor.

\vspace{1cm}

\bibliographystyle{spphys}
\nopagebreak
\bibliography{main.bib}
\end{document}